\documentclass[sn-mathphys-num]{sn-jnl}

\usepackage{multirow}%
\usepackage{amsmath,amssymb,amsfonts}%
\usepackage{booktabs}%

\usepackage{color}
\usepackage{float}
\usepackage{soul}
\usepackage{tikz}
\usetikzlibrary{positioning,fit,calc}
\usepackage{capt-of}

\makeatletter
\newcommand{\mylabel}[2]{#2\def\@currentlabel{#2}\label{#1}}
\makeatother
\DeclareMathOperator*{\E}{\mathbb{E}}
\DeclareMathOperator*{\eff}{\mathit{Eff}}
\newtheorem{definition}{Definition}%

\begin{document}

\title{A framework for measuring the training efficiency of a neural architecture}


\author*[1,2]{\fnm{Eduardo} \sur{Cueto-Mendoza}}\email{eduardo.cuetomendoza@tudublin.ie}

\author[2,3]{\fnm{John} \sur{Kelleher}}\email{john.kelleher@tcd.ie}
\equalcont{These authors contributed equally to this work}

\affil*[1]{\orgdiv{School of Computer Science}, \orgname{TU Dublin}, \orgaddress{\street{Grangegorman}, \city{Dublin 7}, \postcode{D07H6K8}, \state{Co. Dublin}, \country{Ireland}}}

\affil[2]{\orgdiv{ADAPT Research Centre}, \orgname{School of Computer Science and Statistics}, \orgaddress{\street{Trinity College Dublin}, \city{Dublin 2}, \state{Co. Dublin}, \country{Ireland}}}


\abstract{

Measuring Efficiency in neural network system development is an open research problem. This paper presents an experimental framework to measure the training efficiency of a neural architecture. To demonstrate our approach, we analyze the training efficiency of Convolutional Neural Networks and Bayesian equivalents on the MNIST and CIFAR-10 tasks. Our results show that training efficiency decays as training progresses and varies across different stopping criteria for a given neural model and learning task. We also find a non-linear relationship between training stopping criteria, training Efficiency, model size, and training Efficiency.

Furthermore, we illustrate the potential confounding effects of overtraining on measuring the training efficiency of a neural architecture. Regarding relative training efficiency across different architectures, our results indicate that CNNs are more efficient than BCNNs on both datasets. More generally, as a learning task becomes more complex, the relative difference in training efficiency between different architectures becomes more pronounced.

}




\maketitle

\section{Introduction}

Artificial Intelligence is predicted to be a critical enabling technology for many of the 17 Sustainable Development Goals (SDGs). However, its current dependency on massive datasets and computer power means that it will also inhibit the attainment of some SDGs, particularly SDG7 (Affordable and Clean Energy) and SDG 13 (Climate action)~\cite{vinuesa:2020}. Modern Artificial Intelligence (AI) uses data-driven methods like deep learning. It is primarily driven by trends of ever larger datasets, larger models, and more powerful computers with the sole concern of improving model accuracy~\cite{kelleher:2019}. This dynamic resulted in a 300,000x increase between 2012 and 2018 in the computation required to train a competitive DL model~\cite{aiandcompute:2018} (this trend far exceeds Moore's Law). Indeed, it has recently been estimated that training one AI model generated the CO2 emissions equivalent to driving 700,000 km~\cite{deweerdt:2021}.

The environmental challenge posed by AI's growing energy needs and associated carbon emissions has been recognized in recent years. For example, researchers in AI Ethics have highlighted this challenge~\cite{benderetal:2021} and have called for more research on ``sustainable methods of AI''~\cite{wynsberghe:2021}. In response to these calls, there is a growing trend within AI research to move beyond systems evaluations solely based on accuracy. Recent research tends to report hardware details and training time alongside accuracy, and some papers report FLOPS. However, time and FLOPS are not sufficient to characterize Efficiency. There is a growing body of work (e.g.,~\cite{greenai},~\cite{energyandpolicy},~\cite{big},~\cite{gpuproc},~\cite{hard2}) that shows that more data is required to understand the energy and resource trade-off of deep neural networks. Consequently, a critical step in developing sustainable AI is the development of measures for Efficiency that can be integrated into the development process of an AI system.

This paper directly addresses the need for a measure to characterize the Efficiency of a neural network architecture on a specific hardware and learning task. A natural efficiency ratio of interest for a neural architecture is the ratio between the accuracy of a neural model and the energy consumed to achieve this accuracy. Accuracy is usually measured using an appropriate measure for the task and dataset distribution (e.g., F1, AUC-ROC, etc.). However, several recent results highlight a non-linear relationship between the accuracy of a neural model and the size of the model~\cite{descent}. This suggests that there is likely a non-linear relationship between the training efficiency of an architecture and the size of the model instantiating the architecture. At the same time, there is a gap in the research literature in terms of how the training efficiency of a neural architecture varies across training. Understanding the dynamics of training efficiency is crucial as it informs decisions relating to the stopping criterion for training. Consequently, in this work, we set out an experimental methodology for comparing the relative Efficiency of different neural architectures in terms of their efficiency dynamics as training progresses and the changes in Efficiency as the size of the models instantiating the architectures vary. This experimental methodology includes both a measure of Efficiency and an experimental framework for capturing the necessary data for the efficiency measure.

In order to test and demonstrate the usefulness of our efficiency measure, we use our experimental framework to analyze the relative Efficiency of two different neural architectures, a CNN network (LeNet) and a Bayesian Convolutional Network (BCNN), on the MNIST and CIFAR-10 tasks. BCNNs are an interesting case study because, although they have produced better results than LeNet on MNIST and CIFAR-10~\cite{bayes}, their Efficiency relative to standard frequentist networks has yet to be assessed. Furthermore, given that the outcomes obtained by training a frequentist LeNet architecture with backpropagation and its BCNN counterpart trained using approximate variational inference---implemented via dropout---are very different (training the frequentist LeNet results in a point estimate in parameter space, whereas training the BCNN returns a probability distribution over a parameter space), it is likely that there will be differences in terms of Efficiency between these two architectures.

In summary, the key contributions of this research are: (1) we propose a measure of the training efficiency of a neural architecture on a given task; (2) we present a case study analyzing the efficiency dynamics of CNNs and BCNNs on multiple tasks across training; and (3) we analyze the overall Efficiency of CNN versus BCNN architectures. Our results indicate that CNNs are more efficient than BCNNs for training. Also, the Efficiency of both architectures varies across training. For both architectures, there is a non-linear relationship between training efficiency stopping criteria and between training efficiency and model size. Furthermore, we highlight and illustrate the confounding effect that overtraining can have on measuring the Efficiency of a neural architecture. Finally, as the learning task becomes more complex, the relative difference in training efficiency between different architectures becomes more pronounced.

\section{Related Work}\label{sec:related_work}

Research on Efficiency in AI can broadly be categorized into four research streams: architectures, compression, training regimes, and metrics. The first of these streams focuses on developing more computationally efficient neural architectures. For example, improving the Efficiency of the attention mechanism in transformer models~\cite{vaswanietal:2017} has frequently been a target for this type of research. This is due to the popularity of transformer models and the high complexity in time and space $O(n^2)$---of the standard attention mechanism. Within this category of work, the Reformer~\cite{reff} proposes an efficiency improvement (in terms of computation and memory) to the standard transformer that replaces the regular dot-product attention mechanism with one that uses locality-sensitive hashing, and the Linformer~\cite{linf} replaces the transformer attention mechanism and approximates it by a low-rank matrix which reduces the complexity of the attention layer to $O(n)$. A recent survey of work on improving Efficiency in transformers is presented in \cite{TayEtAl-Transformers-2022}. Also, although research on neural architecture search has traditionally focused on optimizing for a single objective (such as accuracy), recently, there has been a growing interest in multi-objective neural architecture search which considers Efficiency (frequently hardware efficiency to enable edge deployment) as part of the optimization problem (see e.g., \cite{zeng2020black, WhiteEtAL-NAS-2023, ChenEtAl-NAS-2023, Lu-MultiObjectiveNAS-2024}).

A second stream of research has focused on improving Efficiency by reducing model size. Some of this work trades extra computation during initial model training for smaller, more efficient models at inference. For example, the EfficientNet~\cite{effn} and EfficientNet v2~\cite{eff2} papers propose model scaling methods that seek to maximize model efficiency during inference (by attempting to minimize the final model depth, width, and resolution) while preserving accuracy at the cost of extra computation during training. Similarly, the training methodology proposed in~\cite{ofa} uses pruning during training to reduce model depth, width, kernel size, and resolution. Another example of this type of work is the  Lottery Ticket Hypothesis~\cite{lott} methodology, which focuses on finding small subnetworks that can fit into different hardware platforms and generalize better. Some research focused on reducing model size is designed to work on pre-trained models. For example, NetAdapt uses empirical measures to reduce several hyperparameters in order to fulfill a certain resource budget~\cite{netada}, and DistilBERT uses model distillation techniques to generate smaller models from a complete BERT transformer~\cite{distb}. \cite{zhou2023optimization} provides a recent review of work on compressing deep neural networks that cover the four main approaches found in the literature (pruning, quantization, factorization, and distillation) and conclude that optimization approaches that combine these different compression approaches are an emerging area of research.

The third stream of research focuses on improving the training regime's Efficiency. Work in this stream generally focuses on modifying one or more of the following components of the training regime: the ordering of (i.e., curriculum learning) or the selection of the training data presented to the model \cite{jiang2019accelerating,mindermann2022prioritized,xie2023data,wang2023efficienttrain,yang2023towards, Wang2024Curriculum}; dynamically modifying the architecture of the model as part of the training process \cite{gong2019efficient,zhang2020accelerating, PanEtAL-Reusing-2023, Ding_2023_CVPR}; modifying the objective function \cite{anil2020scalable,goldfarb2020practical,eschenhagen2024kronecker}; and improving the optimization algorithm \cite{liu2023sophia,chen2024symbolic}.\footnote{We note that within the research on improving optimization algorithms the concept of training efficiency is often framed in terms of the convergence rate achieved by the algorithm for a fixed architecture on a learning task (see e.g. \cite{kingma2014adam,yingetal2024}). By contrast, in this work, we are focused on measuring the training efficiency of a neural architecture (rather than an optimization algorithm) on the task.} \cite{KaddourEtAl-2023} reports a recent empirical study of the effectiveness of several of these efficient training approaches against a baseline training regime that used the Adam optimizer with a fully decayed learning rate. These experiments used a fixed computation budget based on wall time (calculated by multiplying the number of iterations of training by the time per iteration for that architecture and training regime on a reference hardware system) as the criterion for stopping training. Three budgets were used for each experiment: 6 hours, 12 hours, and 24 hours. The results indicate that the tested training modifications did not statistically outperform the baseline in most experiments. When they did, this improvement was reduced as the computing budget increased.

The fourth stream of research is focused on developing measures and methodologies for assessing the performance or Efficiency of an AI solution for a given problem. One focus within this stream of research has been on hardware efficiency, see, e.g., \cite{davis2009flops,gpuproc}. Another focus for this stream of research is on performance or Efficiency during inference. Frequently, this work focuses on pruning models during training to improve Efficiency at inference, see, e.g., \cite{slimming} and \cite{weightconn}, which both use the reduction in floating point operations per inference as a measure of how their pruning approaches improve network efficiency. Examples of work in this area that are relevant to this work include Canziani et al.~\cite{pract}, and  Jurj et al. \cite{jurj2020environmentally}. Both of these works propose measures of Efficiency during inference, and what is particularly relevant for this work is that they use a direct measure of energy consumed (rather than FLOPs) as a measure of resource usage (work done) when calculating Efficiency.
Similarly,~\cite{desislavov2023compute} examines the trends in computational and energy costs associated with deep learning model inference and assesses whether the exponential growth in model parameters translates into a proportional increase in energy consumption. Their analysis considers algorithmic improvements and hardware advancements to understand their impact on energy consumption. We conclude that algorithmic advancements and hardware specialization have significantly improved the energy efficiency of DNNs.

The work most relevant to this research is focused on Efficiency during model training. As noted in~\cite{greenai}, in the research model, training occurs much more frequently than post-deployment inference, so understanding Efficiency during training is in and of itself an important topic. Indeed,~\cite{greenai} reviews several different measures for Efficiency or work done during training (including, \emph{carbon emissions}, \emph{electricity usage}, \emph{elapsed real time}, \emph{number of parameters}, and \emph{floating point operations (FLOPS)}) and argue that FLOPS is the fairest measure to use to compare different approaches. They attribute two properties to FLOPS in support of this argument: (a) FLOPS directly measures the work done when running a specific instance of a model and, therefore, is related to the energy consumed, and (b) it is agnostic to the hardware on which the model is run. However, metrics based on counts of operations performed by a neural network require hardware profiling, and this is computationally expensive to perform~\cite{mills2021profiling}.
Consequently, developing a metric for training efficiency that does not require hardware profiling is desirable. \cite{bartoldson2023compute} presents a recent review of the most commonly used metrics in efficiency research, including training time, FLOPs, number of model parameters, electricity usage, carbon emissions, and operand sizes. Overall, they found that all these metrics have significant limitations in either not directly measuring the factors of interest or being dependent on confounding factors such as hardware, time, etc. Finally, we note that all of the metrics discussed above (be it FLOPs, $CO_2$ emissions \cite{energyandpolicy} or using wall time as a measure \cite{big}) do not consider model accuracy on a task and so do not measure efficiency \emph{per se} but rather are an estimate of work done. We propose a novel efficiency metric considering the relationship between accuracy and work/resource usage.

However, we looked for alternative energy consumption and Efficiency measures during training to avoid the hardware profiling challenges associated with FLOPS measures. Li et al.~\cite{eval} explore the power behavior and energy consumption of several CNN architectures on both CPUs and GPUs, with a particular focus on characterizing the energy consumption of different layer types (convolution, pooling, ReLU, and so on) during training. Similar to Li et al.~\cite{eval} (and  Canziani et al.'s work on inference efficiency~\cite{pract}, and  Strubell et al.'s work on predicting $CO_2$ emissions~\cite{energyandpolicy}), we propose using energy consumed rather than FLOPS as our measure of work done/resource usage. Also, like Canziani et al.~\cite{pract}, we are interested in measuring Efficiency, that is, the relationship between performance (e.g., accuracy) and resource usage (e.g., energy consumed). However, we are focused on the training phase rather than on inference.
Furthermore, like Strubell et al.~\cite{energyandpolicy} and Li et al.~\cite{eval}, we focus on the training phase. However, we go beyond measuring the energy consumed in training a specific model and propose a measure of the relative Efficiency of a neural architecture (distinct from a specific model) on a given task. We compare the LeNet CNN architecture against a Bayesian Convolutional Network (BCNN) as a test case for our efficiency measure. We chose this comparison because BCNNs are not trained with backpropagation, and we conjecture that this comparison may reveal exciting interactions between training regimes and model efficiency.

\section{Defining an efficiency measure for Deep Neural Networks}\label{sec:exp_framework}

\begin{figure}[H]
    \centering

    \begin{tikzpicture}[scale=0.6, every node/.style={transform shape}]
    \tikzset{vertex/.style = {shape=circle,draw,minimum size=1cm}}
    \tikzstyle{arrow} = [thick,->,>={stealth[scale=3]}]
    \Large
        \node[vertex, scale=0.8] (b) at  (0,0) {\;Energy\;};
        \node[vertex, scale=0.8] (c) at  (4,0) {Accuracy};
        \draw [arrow, rounded corners=2, dashed] (b) to (c);
    \end{tikzpicture}

\caption{Network training efficiency visualized as the ratio of accuracy to energy}
\label{fig:acctoenerg}
\end{figure}
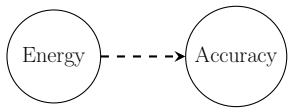

The concept of Efficiency is fundamental to this work:
~\\
\begin{definition}
    Efficiency measures a system's capacity to achieve a goal (measured by a metric) with a given amount of resources.
\end{definition}
~\\
When considering the training efficiency of a neural network on a learning task, it is natural to consider how the accuracy of the network architecture varies as the energy consumed for training changes. This is the efficiency ratio that equation~\ref{def:acctoenergy} defines and that Figure~\ref{fig:acctoenerg} illustrates (in this figure, the arrow represents an efficiency calculation---in the form of Equation \ref{def:acctoenergy}---where the arrow points from the denominator to the numerator).

\begin{equation} \label{def:acctoenergy}
   Efficiency \propto \frac{Accuracy}{Energy}
\end{equation}

However, it is difficult to directly calculate a general estimate of the ratio of energy to accuracy for a given neural architecture on a task because the ratio is dependent on measures used to measure energy and accuracy and is sensitive to hyperparameter decisions (e.g., network size), and training regime decisions (e.g., convergence criteria).
Consequently, in this section, we set out a methodology for calculating this efficiency ratio by averaging across a sequence of experiments that allow for hyperparameters and training regime variations. Then, we used these results to compute our final measure.

\subsection{Metrics for Energy and Accuracy}

Deciding what system components to report energy consumption over is not trivial. For example, although the CPU, GPU, and memory are natural system components to consider when tracking energy consumption during the training of a network, other parts of the system, such as fans, buses, and transistors, also consume energy related to training \cite{gpipe,hard2}. However, due to the difficulties in measuring the energy consumption of these secondary or satellite components, we have decided to focus our analysis on the energy consumed during our experiments from the GPU, CPU, and RAM.

We could use several different measures to measure these components' energy usage. For example, one family of energy measures often used for neural network research is those based on counting the number of computational operations; for example, Schwartz et al.\ suggest using the number of FLOPS~\cite{greenai}. FLOPS, however, is one of many types of operation that can be considered. Data movement operations can be much more expensive regarding energy consumption~\cite{horowitz:2014,szeetal:2020}. However, one of the challenges with tracking energy consumption by counting operations is that the energy consumed by an operation is affected by the sparsity of the data being processed and the data representation being used~\cite{zhengandmazumder:2019}. For example, switching from 32 to 16-bit floating point reduces the energy cost of FLOP operations (and in some cases, this can be done with negligible impact of model accuracy~\cite{mixed}) and also reduces energy consumption by reducing data movement (i.e., reduced memory bandwidth) and reduced energy per memory access (due to smaller memories).

In our experiments\footnote{To demonstrate the applicability of our methodology across different hardware platforms, we replicate the experiments reported in the main body of the paper on different hardware, more details on these experiments are found in~\ref{apx:hardware_results}.} the hardware used was a Tesla T4 with 15109 MiB memory, from Google Colab (driver version 470.63 and CUDA version: 11.2) and energy collection for the GPU was done using NVIDIA System Management Interface version 460.39 and for recording the energy consumed by the CPU and RAM during training we use the powertop\footnote{https://01.org/powertop} system interface which is a Unix native system tool. We used these tools in each experiment to repeatedly sample and record the energy consumption rate by the GPU, CPU, and RAM access components as each network is being trained. We then calculate the Efficiency of the trained model as the ratio between the performance obtained by the model and the total energy consumed~\footnote{measured in terms of Joules per second (Watts)} to train the model, as follows:

\begin{equation}
    \eff(Acc,W,i=epoch) = \frac{Acc_i}{\sum_{n=0}^i[W_n]}
    \label{eq:efficiencyvalidated}
\end{equation}

Where $Acc_i$ is the accuracy obtained on that epoch of training of the model, $\sum[W_n]$ is the sum of the energy samples obtained up to that epoch of training, and $W_n = W_n^{GPU} \bigoplus W_n^{CPU} \bigoplus W_n^{RAM}$, $\bigoplus$ is the concatenation operation.

The selection of the appropriate measure for model performance depends on the task type (e.g., classification, regression, segmentation, and so on) and factors such as the distribution of class labels within the data \cite{fmlpda}. In the experiments we report in this paper, the tasks are classification tasks with balanced label distributions, so we have chosen to use simple accuracy for the task. Specifically, we report a model's accuracy (Acc) on the test set after training has converged. In experiments where we use a hold-out test set methodology, Acc is simply the accuracy of the trained model on the test set. In experiments where we use a $k$ cross-fold validation methodology, Acc is the mean accuracy across the $k$ validation folds.

\begin{figure}[H]
    \centering
    \begin{tikzpicture}[scale=0.6, every node/.style={transform shape}]
    \tikzset{vertex/.style = {shape=circle,draw,minimum size=1cm}}
    \tikzstyle{arrow} = [thick,->,>={stealth[scale=3]}]
    \Large

        \node[vertex, scale=0.8] (b) at  (0,0) {\;Energy\;};
        \node[vertex, scale=0.8] (c) at  (4,0) {Accuracy};

        \draw [arrow, rounded corners=2, dashed] (b) to node[midway,below] {$\scalebox{0.5}[0.5]{\text{Acc}\,/\,\text{W}}$} (c);

    \end{tikzpicture}
\caption{Visualisation of relationships between variables tracked in the experiments}
\label{fig:measuresrelationship}
\end{figure}
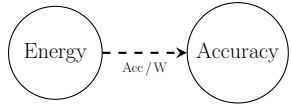

Figure \ref{fig:measuresrelationship} illustrates the relationship between these measures. As seen above, in this figure, the arrows represent efficiency calculation where the arrow points from the denominator to the numerator.
The dashed arrow highlights the overall efficiency calculation we wish to calculate, $\text{Acc}\,/\,\text{W}$, the average amount of task accuracy obtained per unit of energy (Watt) expended in training.

\subsection{Allowing for hyperparameter variations: Model Size}

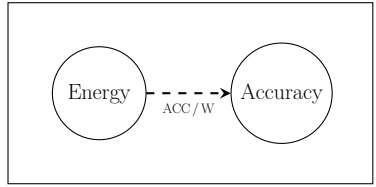
\begin{figure}[H]
    \centering
    \begin{tikzpicture}[scale=0.6, every node/.style={transform shape}]
    \tikzset{vertex/.style = {shape=circle,draw,minimum size=1cm}}
    \tikzstyle{arrow} = [thick,->,>={stealth[scale=3]}]
    \Large
        \node (r1) [draw=black, fill=none, minimum width=8cm,minimum height=4cm]{};
        \node (r1Label)[above=0cm of r1] {\textbf{Model Size: $j{=}\{1,\dots,N\}$}};
        \node[vertex, scale=0.8] (b) at  (-2,0) {\;Energy\;};
        \node[vertex, scale=0.8] (c) at  (2,0) {Accuracy};
        \draw [arrow, rounded corners=2, dashed] (b) to node[midway,below] {$\scalebox{0.5}[0.5]{\text{ACC}\,/\,\text{W}}$} (c);
    \end{tikzpicture}
\caption{Visualisation of how model size is integrated into the experimental methodology}
\label{fig:efficiencybysize}
\end{figure}

To experimentally control for the effect of model size\footnote{We use the term model to denote a particular instantiation of a neural architecture.} we propose to run each experiment multiple times for each neural architecture using a different size model in each run, and for each model size, record both the total energy consumed during training $\sum_{samples}[W]$ and the accuracy obtained by the model.
We then calculate the Efficiency for each model on an experimental task as the ratio of accuracy to the total energy consumed to train it. Finally, we calculate the Efficiency of a neural architecture on an experimental task as the mean Efficiency of the models implementing that architecture on the task. Figure \ref{fig:efficiencybysize} illustrates how model size is included in the experimental design, and Equation \ref{eq:efficiencybysize} defines how we integrate model size into the calculation of the training efficiency of a network architecture.

\begin{equation}
    \eff(arch,j=size) = \E\limits_{n=1}^j\left[\eff(Acc,W,i=epoch)_j\right]
    \label{eq:efficiencybysize}
\end{equation}

\subsection{Training Regime Variations: Convergence Criteria}

The training efficiency of a network (accuracy/energy) is likely to vary as training progresses; in other words, the gain in model accuracy per unit of energy expended is likely to change between the early epochs of training and the later epochs of training. At the same time, the amount of time a network is trained for will vary depending on the convergence criteria used to stop training. To control for this, we define four different convergence criteria and run each experiment with each of these criteria (in combination with our $N$ model size variations, we will run each experiment $N$ times for each of the four convergence criteria). We then calculate the overall training efficiency of network architecture on a task by first calculating the network efficiency for each convergence criterion using Equation~\ref{eq:efficiencybysize} and then calculating the expected value across these efficiency scores.

The four convergence criteria we define are:
\begin{enumerate}
    \item train for a preset number of epochs, in our experiments, we set \textbf{Epochs==50}
    \item train until the model achieves a preset accuracy on a validation set; in our experiments, we set the accuracy target to \textbf{Accuracy==99}
    \item use \textbf{early stopping} as the training convergence method, i.e., we track model accuracy on the validation set across consecutive training epochs. Training stops if accuracy does not increase across a preset number of epochs (known as the \textbf{patience} parameter). In our experiments, we used a level of patience of 3.
    \item stop training after a preset energy (W) budget has been consumed, for our experiments, we set the energy budget to be \textbf{Energy==100kW}
 \end{enumerate}

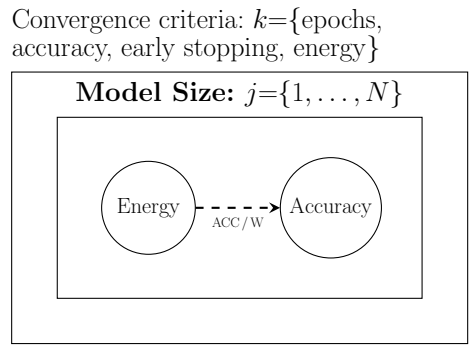
\begin{figure}[H]
    \centering
    \begin{tikzpicture}[scale=0.6, every node/.style={transform shape}]
    \tikzset{vertex/.style = {shape=circle,draw,minimum size=1cm}}
    \tikzstyle{arrow} = [thick,->,>={stealth[scale=3]}]
    \Large
        \node (r0) [draw=black, fill=none, minimum width=10cm,minimum height=6cm]{};
        \node (r0Label)[above=0cm of r0, text width=10cm] {Convergence criteria: $k{=}$\{epochs, accuracy, early stopping, energy\}};
        \node (r1) [draw=black, fill=none, minimum width=8cm,minimum height=4cm]{};
        \node (r1Label)[above=0cm of r1] {\textbf{Model Size: $j{=}\{1,\dots,N\}$}};
        \node[vertex, scale=0.8] (b) at  (-2,0) {\;Energy\;};
        \node[vertex, scale=0.8] (c) at  (2,0) {Accuracy};
        \draw [arrow, rounded corners=2, dashed] (b) to node[midway,below] {$\scalebox{0.5}[0.5]{\text{ACC}\,/\,\text{W}}$} (c);
    \end{tikzpicture}
\caption{Visualisation of how convergence criteria are integrated into the experimental methodology}
\label{fig:efficiencybyconvergence}
\end{figure}

Figure \ref{fig:efficiencybyconvergence} illustrates how these convergence criteria are integrated into the experimental setup, and Equation \ref{eq:meaneffiencybyconvergence} defines how we calculate an overall mean training efficiency for a network architecture that accounts for both model size and convergence criteria.

\begin{equation}
    \eff(arch,k=convergence) = \E\limits_k\left[\eff(arch,j=size)_k\right]
    \label{eq:meaneffiencybyconvergence}
\end{equation}
Where in the case of Equation~\ref{eq:meaneffiencybyconvergence}, $\eff(arch, size)_k$ is computed as in Equation~\ref{eq:efficiencybysize}.

\section{Case Study: Convolutional and Bayesian Convolutional Architectures}\label{sec:casestudy}

In this case study, we demonstrate the use of our efficiency framework by comparing the Efficiency of a CNN network (LeNet) with that of a Bayesian Convolutional Network (BCNN).
The BCNN network is trained using approximate variational inference, which is implemented via dropout.
Similar to the experiments reported in the original BCNN paper~\cite{bayes}, we use the LeNet-5 architecture from~\cite{lecunetal:1998} as the baseline architecture for our experiments. Following~\cite{bayes}, the corresponding Bayesian version of the LeNet baseline was created by applying a dropout with a probability of 0.5 after all convolution and weight layers (i.e., this is the model called ``lenet-all'' in~\cite{bayes}). Tables \ref{tab:lenet} and \ref{tab:bcnn} report the hyperparameters used to train the LeNet and BCNN models (note: we use the same hyper-parameter settings as reported for experiments performed by~\cite{bayes}).

\begin{table}[h!]
    \begin{minipage}{.5\linewidth}
      \centering
        \caption{LeNet hyperparameters}
        \label{tab:lenet}
        \begin{tabular}{l l}
            \hline
            Architecture & LeNet-5 \\
            ~ & ~\\
            \hline
            epochs & 50 \\
            learning rate & 0.001 \\
            num workers & 4 \\
            batch size & 256 \\
            activation & soft plus \\
            loss & cross-entropy \\
            optimiser & ADAM \\
            initialization & Normal (mean:0, \\ 
            & variance:1) \\
            \hline
        \end{tabular}


    \end{minipage}%
    \begin{minipage}{.5\linewidth}
      \centering
        \caption{BCNN hyperparameters}
        \label{tab:bcnn}
       \begin{tabular}{l l}
        \hline
        Architecture & LeNet-5 \\
        & (Bayesian filters) \\
        \hline
        epochs & 50 \\
        learning rate & 0.001 \\
        num workers & 4 \\
        batch size & 256 \\
        activation & soft plus \\
        loss & cross-entropy \\
        optimiser & ADAM  \\
       sample size & $10^{-25}$ \\
        train ensemble & 1 \\
        test ensemble & 1 \\
        $\beta$ & 0.1 \\
        prior $\mu$ & 0.0 \\
        prior $\sigma$ & 0.1 \\
        posterior $\mu_{init}$ & (0,0.1),\\
        & (mean, std) \\
        posterior $\rho_{init}$ & (-5,0.1), \\
        \hline
    \end{tabular}

    \end{minipage}
\end{table}

The two models described above are baseline versions of the models used in our experiments. However, in each of our experiments, we vary the model size and different convergence criteria to explore and contrast the efficiency trade-offs for each architecture between size and accuracy and size and Efficiency. In the Bayesian case, two ways of approximating the posterior probability distribution exist: Variational Inference (VI) and Markov Chain Monte Carlo (MCMC). In most cases, (VI) performs excellently but is not a great estimator. While MCMC can be computationally expensive but is an excellent estimator \cite{bnn1}, in our experiments, we estimate the posterior using Variational Inference.
The best strategy (depth versus width) for scaling a model is an open research challenge. However, because both architectures we consider here are convolutional networks, we decided to scale the models by increasing the filters used in each layer. In other words, we scaled the width of the models, and we did this by multiplying the number of filters in each layer by multiples from $\times1$ up to $\times5$ the original baseline size. This means that in our experiments, we test five versions of the LeNet architecture: LeNet-1, the baseline architecture is the same as reported in~\cite{lecunetal:1998}, LeNet-2 has twice the number of filters in each layer as LeNet-1, LeNet-3 has three times the number of filters, and so on up to LeNet-5 with five times the number of filters. Similarly, BCNN-1 is the baseline Bayesian architecture from~\cite{bayes} and has the same size as LeNet-1, and BCNN-2 through BCNN-5 are scaled to match their corresponding LeNet-X counterparts in size and structure.

All four efficiency experiments were performed on the MNIST \cite{mnist} and CIFAR-10 \cite{cifar} datasets. The same hyperparameters were used for the architectures on both the MNIST and CIFAR-10 datasets\footnote{https://unix-talk.com/TastyPancakes/bayesiancnn.git, has all the author's code for the experiments.}. Both of these datasets are based on the task of handwritten numeric digit recognition images, with each image containing a single handwritten digit between 0 and 9. The MNIST dataset contains 10,000 images across ten classes (0-9), each being a $28 \times 28$ pixel gray-scale image. The CIFAR-10 also has ten classes with 6000 images per class, each color image being $32 \times 32$ pixels. The original experiments with MNIST and CIFAR-10 used different experimental methods: MNIST used a single training and test split, whereas CIFAR-10 used a six-fold cross-validation methodology. In our experiments, we follow the same experimental methodology for each dataset as was reported in the original experiments.
Consequently, for the MNIST dataset, a simple split was performed with the training set consisting of 60,000 handwritten digits and our test set of 10,000, and the label distributions in both the training and test sets are balanced across all ten digits. So, in all of our experiments, when we report an accuracy on the MNIST data, this is the accuracy obtained by the model on the single hold-out test set. By contrast, for the CIFAR-10 dataset, we use a 6-fold cross-validation methodology in each experiment, where each fold contains exactly 1000 randomly selected images from each class, and the reported accuracy is the average accuracy of an architecture across these folds.

\subsection{Results from the case study} \label{sec:measure_results}

This section presents the results for the 50 epoch, early-stopping, energy-bound, and accuracy-bound experiments. For each experiment, dataset, and neural architecture, we present a table showing the efficiency calculation across model size for each architecture under the convergence criteria specified in that experiment (using Equation \ref{eq:efficiencybysize}). Note that in~\href{https://osf.io/qw7rj/?view_only=56f77a9e6a8245048b2531c9d3a076b0}{supplementary material} we present, for each experiment, plots of the training and test accuracy by training epoch for each model.\footnote{All data in the tables from Section~\ref{sec:casestudy}, and Section~\ref{sec:analysingflops}, is released at: \href{https://osf.io/2jd85/?view\_only=9abd36ef80b6467eabaaed3bcb983031}{Open Science Foundation}}

\subsubsection{50 epoch experiment}
\label{subsubsec:100epoch}

In this first experiment, the stopping criterion for training was set at 50 epochs. For each architecture (LeNet and BCNN), the experiment is run a total of 10 times per architecture: once for each of the 5 model sizes  (LeNet-1 to LeNet-5, and BCNN-1 to BCNN-5) on both the datasets (MNIST and CIFAR). During each run of the experiment, we repeatedly recorded the energy being consumed and the amount of memory (GPU and RAM) being used (recorded as model size (MiB) size in RAM and GPU memory).

Table~\ref{tab:mnist_100_eff} and Table~\ref{tab:cifar_100_eff} show the efficiency calculation using a convergence criterion of 50 epochs. Note that for the CIFAR dataset, we use a six-fold cross-validation methodology. So for this dataset, the accuracy reported for each model size $i$ ($Acc_i$) in Table \ref{tab:cifar_100_eff} is the average accuracy for that model size across the six validation sets after training has converged.

\begin{table}[h!]
    \centering
    \caption{MNIST compute for the 50 epochs experiment.}
    \label{tab:mnist_100_eff}
    \begin{tabular}{c | c c c c c}
         Model & Epochs & $Acc_i$ & $\sum_{samples}[W]$ & $\eff(Acc,W,epoch)$ & $\eff(arch,size)$\\
         \hline
         BCNN-1 & 50 & $0.97 \times 10^{-1}$ & $2.08 \times 10^{5}$ & $4.67 \times 10^{-6}$ & \multirow{5}{*}{$2.59 \times 10^{-6}$}\\
         BCNN-2 & 50 & $9.78 \times 10^{-1}$ & $4.18 \times 10^{5}$ & $2.34 \times 10^{-6}$ & \\
         BCNN-3 & 50 & $9.77 \times 10^{-1}$ & $4.64 \times 10^{5}$ & $2.11 \times 10^{-6}$ & \\
         BCNN-4 & 50 & $9.77 \times 10^{-1}$ & $5.00 \times 10^{5}$ & $1.95 \times 10^{-6}$ & \\
         BCNN-5 & 50 & $9.76 \times 10^{-1}$ & $5.14 \times 10^{5}$ & $1.90 \times 10^{-6}$ & \\
         \hline
         LeNet-1 & 50 & $9.91 \times 10^{-1}$ & $0.97 \times 10^{5}$ & $10.15 \times 10^{-6}$ & \multirow{5}{*}{$8.09 \times 10^{-6}$}\\
         LeNet-2 & 50 & $9.93 \times 10^{-1}$ & $0.90 \times 10^{5}$ & $11.02 \times 10^{-6}$ & \\
         LeNet-3 & 50 & $9.94 \times 10^{-1}$ & $1.45 \times 10^{5}$ & $6.83 \times 10^{-6}$ &\\
         LeNet-4 & 50 & $9.95 \times 10^{-1}$ & $1.52 \times 10^{5}$ & $6.52 \times 10^{-6}$ &\\
         LeNet-5 & 50 & $9.94 \times 10^{-1}$ & $1.67 \times 10^{5}$ & $5.94 \times 10^{-6}$ &\\
        \hline
    \end{tabular}

\end{table}

\begin{table}[h!]
  \centering
    \caption{CIFAR compute for the 50 epochs experiment.}
    \label{tab:cifar_100_eff}
  \begin{tabular}{c | c c c c c}
    Model & Epochs & $Acc_i$ & $\sum_{samples}[W]$ & $\eff(Acc,W,epoch)$ & $\eff(arch,size)$\\
    \hline
    BCNN-1 & 50 &  $4.35 \times 10^{-1}$ & $2.87 \times 10^{5}$ & $1.51 \times 10^{-6}$ & \multirow{5}{*}{$1.54 \times 10^{-6}$}\\
    BCNN-2 & 50 &  $4.93 \times 10^{-1}$ & $3.09 \times 10^{5}$ & $1.59 \times 10^{-6}$ & \\
    BCNN-3 & 50 &  $5.12 \times 10^{-1}$ & $3.29 \times 10^{5}$ & $1.55 \times 10^{-6}$ & \\
    BCNN-4 & 50 &  $5.24 \times 10^{-1}$ & $3.41 \times 10^{5}$ & $1.54 \times 10^{-6}$ & \\
    BCNN-5 & 50 &  $5.24 \times 10^{-1}$ & $3.44 \times 10^{5}$ & $1.52 \times 10^{-6}$ & \\
    \hline
    LeNet-1 & 50 & $6.35 \times 10^{-1}$ & $0.78 \times 10^{5}$ & $8.13 \times 10^{-6}$ & \multirow{5}{*}{$7.78 \times 10^{-6}$}\\
    LeNet-2 & 50 & $7.18 \times 10^{-1}$ & $0.84 \times 10^{5}$ & $8.47 \times 10^{-6}$ & \\
    LeNet-3 & 50 & $7.71 \times 10^{-1}$ & $0.89 \times 10^{5}$ & $8.66 \times 10^{-6}$ & \\
    LeNet-4 & 50 & $7.88 \times 10^{-1}$ & $0.99 \times 10^{5}$ & $7.91 \times 10^{-6}$ & \\
    LeNet-5 & 50 & $7.96 \times 10^{-1}$ & $1.39 \times 10^{5}$ & $5.72 \times 10^{-6}$ & \\
    \hline
  \end{tabular}

\end{table}

\subsubsection{Early-Stopping experiment}
\label{subsubsec:early}

This experiment has the same design as the 50 epoch experiment presented above, with a single change in the convergence criteria used for training; in this experiment, we use early-stopping criteria for accuracy.

For the MNIST dataset, Table \ref{tab:mnist_es_eff} lists the efficiency calculation using Equation \ref{eq:efficiencybysize}. For CIFAR Table \ref{tab:cifar_es_eff} presents the efficiency calculation using Equation \ref{eq:efficiencybysize}.

\begin{table}[h!]
  \centering
    \caption{MNIST compute for the early-stopping experiment.}
    \label{tab:mnist_es_eff}
  \begin{tabular}{c | c c c c c}
    Model & Epochs & $Acc_i$ & $\sum_{samples}[W]$ & $\eff(Acc,W,epoch)$ & $\eff(arch,size)$\\
    \hline
    BCNN-1 & 65 &  $9.67 \times 10^{-1}$ & $9.19 \times 10^{5}$ & $1.05 \times 10^{-6}$ & \multirow{5}{*}{$1.00 \times 10^{-6}$}\\
    BCNN-2 & 21 &  $9.42 \times 10^{-1}$ & $6.31 \times 10^{5}$ & $1.49 \times 10^{-6}$ & \\
    BCNN-3 & 37 &  $9.61 \times 10^{-1}$ & $9.26 \times 10^{5}$ & $1.04 \times 10^{-6}$ & \\
    BCNN-4 & 53 &  $9.64 \times 10^{-1}$ & $12.21 \times 10^{5}$ & $0.79 \times 10^{-6}$ & \\
    BCNN-5 & 65 &  $9.67 \times 10^{-1}$ & $15.40 \times 10^{5}$ & $0.63 \times 10^{-6}$ & \\
    \hline
    LeNet-1 & 16 & $9.75 \times 10^{-1}$ & $0.75 \times 10^{5}$ & $12.83 \times 10^{-6}$ & \multirow{5}{*}{$8.77 \times 10^{-6}$}\\
    LeNet-2 & 12 & $9.79 \times 10^{-1}$ & $0.59 \times 10^{5}$ & $16.40 \times 10^{-6}$ & \\
    LeNet-3 & 56 & $9.93 \times 10^{-1}$ & $2.69 \times 10^{5}$ & $3.68 \times 10^{-6}$ & \\
    LeNet-4 & 28 & $9.91 \times 10^{-1}$ & $2.12 \times 10^{5}$ & $4.65 \times 10^{-6}$ & \\
    LeNet-5 & 20 & $9.90 \times 10^{-1}$ & $1.58 \times 10^{5}$ & $6.26 \times 10^{-6}$ & \\
    \hline
  \end{tabular}

\end{table}

\begin{table}[h!]
    \centering
    \caption{CIFAR compute for the early-stopping experiment.}
    \label{tab:cifar_es_eff}
    \begin{tabular}{c | c c c c c}
    Model & Epochs & $Acc_i$ & $\sum_{samples}[W]$ & $\eff(Acc,W,epoch)$ & $\eff(arch,size)$\\
         \hline
         BCNN-1 & 61 & $4.39 \times 10^{-1}$ & $3.58 \times 10^{5}$ & $1.23 \times 10^{-6}$ & \multirow{5}{*}{$1.02 \times 10^{-6}$}\\
         BCNN-2 & 41 & $4.23 \times 10^{-1}$ & $3.68 \times 10^{5}$ & $1.15 \times 10^{-6}$ & \\
         BCNN-3 & 41 & $4.40 \times 10^{-1}$ & $4.66 \times 10^{5}$ & $0.94 \times 10^{-6}$ & \\
         BCNN-4 & 21 & $3.86 \times 10^{-1}$ & $2.94 \times 10^{5}$ & $1.31 \times 10^{-6}$ & \\
         BCNN-5 & 81 & $4.92 \times 10^{-1}$ & $1.068 \times 10^{5}$ & $0.46 \times 10^{-6}$ & \\
         \hline
         LeNet-1 & 56 & $5.93 \times 10^{-1}$ & $1.64 \times 10^{5}$ & $3.60 \times 10^{-6}$ & \multirow{5}{*}{$4.93 \times 10^{-6}$}\\
         LeNet-2 & 40 & $6.53 \times 10^{-1}$ & $1.67 \times 10^{5}$ & $3.90 \times 10^{-6}$ & \\
         LeNet-3 & 24 & $6.45 \times 10^{-1}$ & $0.92 \times 10^{5}$ & $6.96 \times 10^{-6}$ & \\
         LeNet-4 & 24 & $6.65 \times 10^{-1}$ & $1.18 \times 10^{5}$ & $5.63 \times 10^{-6}$ & \\
         LeNet-5 & 28 & $7.18 \times 10^{-1}$ & $1.57 \times 10^{5}$ & $4.56 \times 10^{-6}$ & \\
        \hline
    \end{tabular}

\end{table}

\subsubsection{Energy bound experiment}
\label{subsubsec:energy}

In this experiment, the convergence criterion used to stop training was when the energy samples recorded for a training run on an architecture cumulatively summed up to 100,000 W. Apart from this, the design of the experiment is the same as those reported in the previous two sections.

Mirroring the results from the previous experiments, for the MNIST dataset, Table \ref{tab:mnist_eb_eff} lists the efficiency calculation using Equation \ref{eq:efficiencybysize}. Similarly, for CIFAR, Table \ref{tab:cifar_eb_eff} presents the efficiency calculation using Equation \ref{eq:efficiencybysize}. Note that some of the values for total energy listed in the results for this experiment are above the training convergence criterion of 100,000W. These values are correct values from the experiment. The reason for these values is that although we sample throughout the training process (the average sampling rate for energy was 973 per second for the NVIDIA system and 1052 samples per second for the AMD system), we perform the check of the cumulative amount of energy consumed during training at the end of each epoch. Consequently, the energy consumed during a training run exceeds the stropping threshold if the process crosses that threshold during an epoch. 

\begin{table}[h!]
    \centering
    \caption{MNIST compute for the energy bound experiment.}
    \label{tab:mnist_eb_eff}
    \begin{tabular}{c | c c c c c}
    Model & Epochs & $Acc_i$ & $\sum_{samples}[W]$ & $\eff(Acc,W,epoch)$ & $\eff(arch,size)$\\
         \hline
    BCNN-1 & 19 &  $9.44 \times 10^{-1}$ & $1.46 \times 10^{5}$ & $6.43 \times 10^{-6}$ & \multirow{5}{*}{$6.18 \times 10^{-6}$}\\
    BCNN-2 & 13 &  $9.33 \times 10^{-1}$ & $1.62 \times 10^{5}$ & $5.73 \times 10^{-6}$ & \\
    BCNN-3 & 10 &  $9.19 \times 10^{-1}$ & $1.57 \times 10^{5}$ & $5.83 \times 10^{-6}$ & \\
    BCNN-4 & 08 &  $9.06 \times 10^{-1}$ & $1.43 \times 10^{5}$ & $6.32 \times 10^{-6}$ & \\
    BCNN-5 & 06 &  $8.83 \times 10^{-1}$ & $1.33 \times 10^{5}$ & $6.60 \times 10^{-6}$ & \\
         \hline
    LeNet-1 & 43 & $9.87 \times 10^{-1}$ & $1.16 \times 10^{5}$ & $8.47 \times 10^{-6}$ & \multirow{5}{*}{$8.48 \times 10^{-6}$}\\
    LeNet-2 & 39 & $9.90 \times 10^{-1}$ & $1.16 \times 10^{5}$ & $8.46 \times 10^{-6}$ & \\
    LeNet-3 & 27 & $9.89 \times 10^{-1}$ & $1.15 \times 10^{5}$ & $8.58 \times 10^{-6}$ & \\
    LeNet-4 & 23 & $9.89 \times 10^{-1}$ & $1.15 \times 10^{5}$ & $8.60 \times 10^{-6}$ & \\
    LeNet-5 & 21 & $9.89 \times 10^{-1}$ & $1.19 \times 10^{5}$ & $8.29 \times 10^{-6}$ & \\
        \hline
    \end{tabular}

\end{table}

\begin{table}[h!]
    \centering
    \caption{CIFAR compute for the energy bound experiment.}
    \label{tab:cifar_eb_eff}
    \begin{tabular}{c | c c c c c}
    Model & Epochs & $Acc_i$ & $\sum_{samples}[W]$ & $\eff(Acc,W,epoch)$ & $\eff(arch,size)$\\
         \hline
         BCNN-1 & 19 &  $1.40 \times 10^{-1}$ & $1.49 \times 10^{5}$ & $0.94 \times 10^{-6}$ & \multirow{5}{*}{$2.21 \times 10^{-6}$}\\
         BCNN-2 & 14 &  $3.59 \times 10^{-1}$ & $1.36 \times 10^{5}$ & $2.64 \times 10^{-6}$ & \\
         BCNN-3 & 11 &  $3.25 \times 10^{-1}$ & $1.14 \times 10^{5}$ & $2.85 \times 10^{-6}$ & \\
         BCNN-4 & 09 &  $3.07 \times 10^{-1}$ & $1.31 \times 10^{5}$ & $2.33 \times 10^{-6}$ & \\
         BCNN-5 & 07 &  $2.71 \times 10^{-1}$ & $1.19 \times 10^{5}$ & $2.28 \times 10^{-6}$ & \\
         \hline
         LeNet-1 & 39 & $5.72 \times 10^{-1}$ & $1.18 \times 10^{5}$ & $4.83 \times 10^{-6}$ & \multirow{5}{*}{$5.17 \times 10^{-6}$}\\
         LeNet-2 & 36 & $6.44 \times 10^{-1}$ & $1.15 \times 10^{5}$ & $5.59 \times 10^{-6}$ & \\
         LeNet-3 & 32 & $6.77 \times 10^{-1}$ & $1.16 \times 10^{5}$ & $5.83 \times 10^{-6}$ & \\
         LeNet-4 & 25 & $6.80 \times 10^{-1}$ & $1.17 \times 10^{5}$ & $5.77 \times 10^{-6}$ & \\
         LeNet-5 & 21 & $6.71 \times 10^{-1}$ & $1.75 \times 10^{5}$ & $3.83 \times 10^{-6}$ & \\
        \hline
    \end{tabular}

\end{table}

\subsubsection{Accuracy bound experiment}
\label{subsubsec:accuracy}

The convergence criteria used in these experiments was to stop training when a model obtained a specified accuracy threshold. For the MNIST dataset this accuracy threshold was set at $99\%$ on the training set, and on the CIFAR dataset (where we used a six-fold cross-validation methodology) for each fold the training was stopped when the model had obtained an accuracy threshold of $50\%$ on the training data for that fold\footnote{See the final paragraph of Section \ref{sec:casestudy} for details of the training and test split used for MNIST and the six-fold cross-validation methodology used for CIFAR.}. Our reason for using a lower accuracy threshold for CIFAR was that an accuracy threshold $>50\%$ required training to proceed for more time than our Collab account allowed, and if this time threshold was exceeded, then the training was interrupted, and results were lost.

For MNIST Table \ref{tab:mnist_ab_eff} lists the efficiency calculation using Equation \ref{eq:efficiencybysize}. Similarly, for CIFAR, Table \ref{tab:cifar_ab_eff} presents the efficiency calculation using Equation \ref{eq:efficiencybysize}.

\begin{table}[h!]
    \centering
    \caption{MNIST compute for the accuracy bound experiment.}
    \label{tab:mnist_ab_eff}
    \begin{tabular}{c | c c c c c}
    Model & Epochs & $Acc_i$ & $\sum_{samples}[W]$ & $\eff(Acc,W,epoch)$ & $\eff(arch,size)$\\
         \hline
    BCNN-1 & 72 &  $9.70 \times 10^{-1}$ & $13.32 \times 10^{5}$ & $0.73 \times 10^{-6}$ & \multirow{5}{*}{$1.74 \times 10^{-6}$}\\
    BCNN-2 & 69 &  $9.71 \times 10^{-1}$ & $14.05 \times 10^{5}$ & $0.69 \times 10^{-6}$ & \\
    BCNN-3 & 69 &  $9.70 \times 10^{-1}$ & $14.81 \times 10^{5}$ & $0.66 \times 10^{-6}$ & \\
    BCNN-4 & 77 &  $9.72 \times 10^{-1}$ & $1.60 \times 10^{5}$ & $6.06 \times 10^{-6}$ & \\
    BCNN-5 & 80 &  $9.72 \times 10^{-1}$ & $17.06 \times 10^{5}$ & $0.57 \times 10^{-6}$ & \\
         \hline
    LeNet-1 & 12 & $9.70 \times 10^{-1}$ & $0.57 \times 10^{5}$ & $16.86 \times 10^{-6}$ & \multirow{5}{*}{$26.10 \times 10^{-6}$}\\
    LeNet-2 & 08 & $9.73 \times 10^{-1}$ & $0.44 \times 10^{5}$ & $22.02 \times 10^{-6}$ & \\
    LeNet-3 & 06 & $9.74 \times 10^{-1}$ & $0.36 \times 10^{5}$ & $26.73 \times 10^{-6}$ & \\
    LeNet-4 & 06 & $9.75 \times 10^{-1}$ & $0.36 \times 10^{5}$ & $26.47 \times 10^{-6}$ & \\
    LeNet-5 & 04 & $9.75 \times 10^{-1}$ & $0.25 \times 10^{5}$ & $38.44 \times 10^{-6}$ & \\
        \hline
    \end{tabular}

\end{table}

\begin{table}[h!]
    \centering
    \caption{CIFAR compute for the accuracy bound experiment.}
    \label{tab:cifar_ab_eff}
    \begin{tabular}{c | c c c c c}
    Model & Epochs & $Acc_i$ & $\sum_{samples}[W]$ & $\eff(Acc,W,epoch)$ & $\eff(arch,size)$\\
         \hline
         BCNN-1 & 51 & $4.24 \times 10^{-1}$ & $5.78 \times 10^{5}$ & $0.73 \times 10^{-6}$ & \multirow{5}{*}{$0.68 \times 10^{-6}$}\\
         BCNN-2 & 37 & $4.17 \times 10^{-1}$ & $4.98 \times 10^{5}$ & $0.84 \times 10^{-6}$ & \\
         BCNN-3 & 30 & $4.14 \times 10^{-1}$ & $5.32 \times 10^{5}$ & $0.78 \times 10^{-6}$ & \\
         BCNN-4 & 36 & $4.21 \times 10^{-1}$ & $8.07 \times 10^{5}$ & $0.52 \times 10^{-6}$ & \\
         BCNN-5 & 33 & $4.24 \times 10^{-1}$ & $8.32 \times 10^{5}$ & $0.51 \times 10^{-6}$ & \\
         \hline
         LeNet-1 & 7 & $4.22 \times 10^{-1}$ & $0.60 \times 10^{5}$ & $7.00 \times 10^{-6}$ & \multirow{5}{*}{$10.35 \times 10^{-6}$}\\
         LeNet-2 & 4 & $4.29 \times 10^{-1}$ & $0.37 \times 10^{5}$ & $11.55 \times 10^{-6}$ & \\
         LeNet-3 & 4 & $4.63 \times 10^{-1}$ & $0.43 \times 10^{5}$ & $10.65 \times 10^{-6}$ & \\
         LeNet-4 & 3 & $4.42 \times 10^{-1}$ & $0.33 \times 10^{5}$ & $13.24 \times 10^{-6}$ & \\
         LeNet-5 & 3 & $4.68 \times 10^{-1}$ & $0.50 \times 10^{5}$ & $9.30 \times 10^{-6}$ & \\
        \hline
    \end{tabular}
\end{table}

\section{Analysis of experimental data}\label{sec:analysingflops}

This section presents the analysis of the data obtained from our experiments regarding how Efficiency behaves as training progresses, the relationship between model size and Efficiency, and the relative overall Efficiency of the LeNet and BCNN architectures.

\subsection{Efficiency as training progresses}
\label{secsub:efficiencyacrosstraining}

\begin{figure}[h!]
    \begin{center}
        \includegraphics[width=0.95\textwidth]{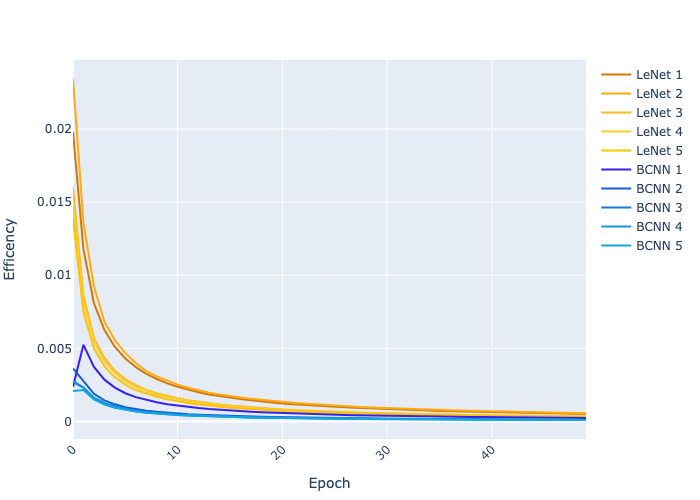}
    \end{center}
    \caption{Efficiency per epoch (MNIST dataset) of the 50 epoch experiment.}\label{fig:mnist_eff_epo}
\end{figure}

\begin{figure}[h!]
    \begin{center}
        \includegraphics[width=0.95\textwidth]{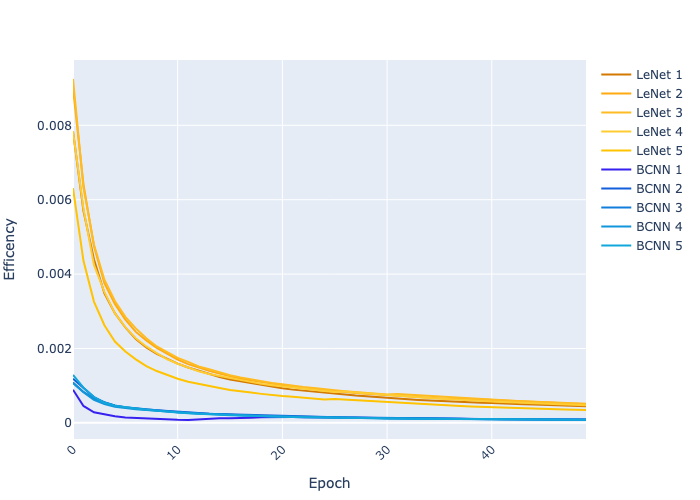}
    \end{center}
    \caption{Efficiency per epoch (CIFAR dataset) of the 50 epoch experiment.}\label{fig:cifar_eff_epo}
\end{figure}

Figure~\ref{fig:mnist_eff_epo} and Figure~\ref{fig:cifar_eff_epo} plot for each of the models trained (LeNet sizes 1--5, and BCNN sizes 1--5) how the Efficiency of the model changes across epochs as training progresses. We base this analysis solely on the results from the 50 epoch experiment because, in this experiment, we have collected the same number of epochs for all sizes and both architectures. As a result, the x-axis, which records the training epochs, goes from 0 to 50 in both figures. The y-axis in the graph plots the Efficiency of a model at a given epoch as defined by Equation~\ref{eq:efficiencyvalidated}. This definition of \textit{Efficiency} is the ratio of a model's performance on a validation set after epoch $i$ of training to the cumulative energy expended in training the model up to that point in training.

For these results, we observe that Efficiency decreases as time progresses. These plots show that although we would expect the performance of a model to improve as training progresses, the rate of improvement tends to decrease as training progresses. After a certain amount of training (epochs), performance plateaus and further training result in energy being expended. Notice that the plots in Figure~\ref{fig:mnist_eff_epo} drop more steeply than those in Figure~\ref{fig:cifar_eff_epo}. This reflects the fact that on the more straightforward MNIST dataset, model performance saturates very early on, whereas, on the more complex CIFAR dataset, it takes more epochs for the models to reach this performance saturation point.

The relative difficulty of the two datasets is also reflected in the differences in the y-axis scales between Figure~\ref{fig:mnist_eff_epo} and Figure~\ref{fig:cifar_eff_epo}. The maximum Efficiency recorded for any models at any epoch on MNIST is above 0.02, whereas on CIFAR, it is below 0.01. The primary driver of this difference is that on MNIST, the models achieved accuracies of 0.97--0.99 (see Table~\ref{tab:mnist_100_eff}), whereas on CIFAR, the range of accuracies of the BCNN models is 0.43--0.53 and the LeNet models 0.66--0.80 (see Table~\ref{tab:cifar_100_eff}).

Finally, comparing Figure~\ref{fig:mnist_eff_epo} with Figure~\ref{fig:cifar_eff_epo}, it is apparent that the gap between the plots for the LeNet models and the BCNN models is more significant in Figure~\ref{fig:cifar_eff_epo}. This suggests that as a learning task becomes more complex, differences in Efficiency become more pronounced.

\subsection{Relationship between stopping criteria and Efficiency, and model size and Efficiency}
\label{ssec:sizeandefficiency}

The results presented in Tables ~\ref{tab:mnist_100_eff}--~\ref{tab:cifar_ab_eff} reveal significant variation in architecture efficiency across different stopping criteria.  Note that this analysis considers the variation in Efficiency by model size. This variation is particularly noticeable in the MNIST dataset. Table~\ref{tab:mnist_eff_bystopping} summarises (from Tables~\ref{tab:mnist_100_eff},~\ref{tab:mnist_es_eff},~\ref{tab:mnist_eb_eff} and~\ref{tab:mnist_ab_eff}) the efficiency results for both architectures across the four stopping criteria on the MNIST dataset. Examining the results for LeNet, the maximum Efficiency ($0.00002610$) is obtained using an accuracy bound stopping criterion, and the minimum Efficiency ($0.00000809$) is recorded using the 50 epoch criterion. This means that LeNet is, averaging across model sizes, approximately $3.22$ times more efficient on MNIST when the accuracy bound criterion is applied compared to the 50 epoch criterion. A similar variation in Efficiency across stopping criteria is observable for the BCNN architecture. However, the criteria that result in the maximum and minimum values differ. For the BCNN architecture on MNIST, using an energy bound stopping criterion gives the maximum Efficiency of $0.00000618$ compared to the minimum Efficiency of $0.00000100$ using early stopping, a variation in Efficiency of $6.18$ times. More generally, we observe a complex non-linear interaction across architectures and convergence criteria, as shown in Figure~\ref{fig:mnist-stopping-eff}, which plots the LeNet versus BCNN efficiency scores by convergence criteria. The within-architecture efficiency variation across stopping criteria and the complex interactions across architectures and stopping criteria highlight the need to include multiple stopping criteria within the efficiency framework.

\begin{minipage}{\textwidth}
 ~\\
    \begin{minipage}[b]{0.40\textwidth}
    \centering
	\begin{tabular}{lrr}
	\hline
	~ & LeNet & BCNN\\
	\hline
    50 Epoch  &	$8.09 \times 10^{-6}$ & $2.59 \times 10^{-6}$ \\
	Early Stopping  &	$8.77 \times 10^{-6}$ & $1.00 \times 10^{-6}$ \\
	Energy Bounded 	 & $8.48 \times 10^{-6}$ & $6.18 \times 10^{-6}$ \\
	Acc. Bounded  &	$26.10 \times 10^{-6}$ & $1.74 \times 10^{-6}$ \\
	\hline
     \end{tabular}
      \captionof{table}{MNIST mean efficiency scores for LeNet and BCNN by stopping criteria}
 \label{tab:mnist_eff_bystopping}
    \end{minipage}
  \hfill
  \begin{minipage}[b]{0.45\textwidth}
    \centering
    \includegraphics[clip=true, trim={100 135 80 155}, width=0.8\linewidth]{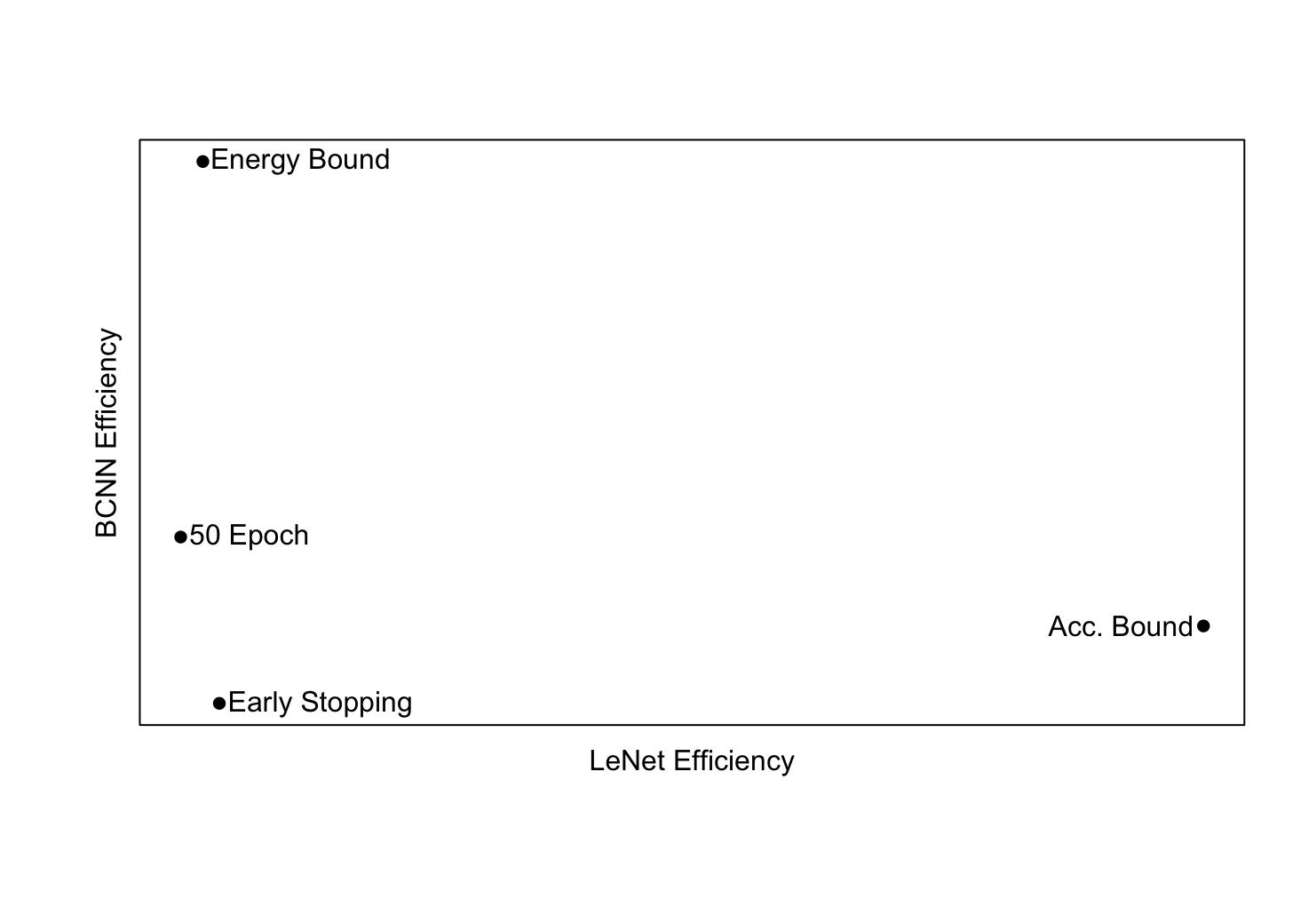}
    \captionof{figure}{LeNet versus BCNN efficiency on MNIST by stopping criteria}
    \label{fig:mnist-stopping-eff}
  \end{minipage}
  ~\\
  \end{minipage}

\begin{figure}[h!]
    \begin{center}
        \includegraphics[width=0.95\textwidth]{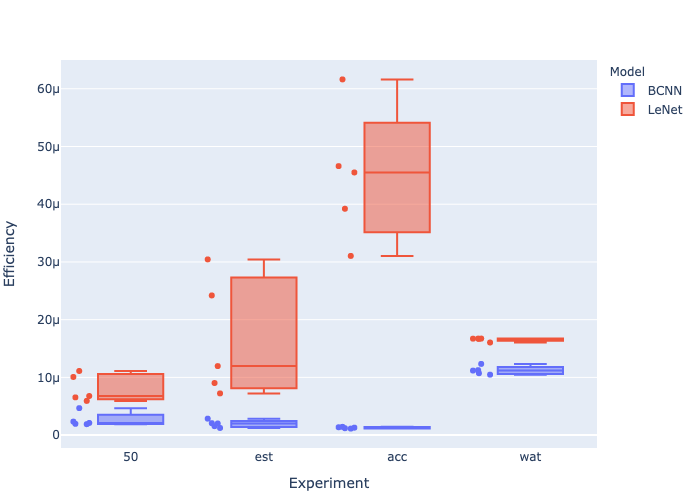}
    \end{center}
    \caption{Efficiency for the 4 experiments (MNIST dataset).}\label{fig:mnist_efficiency}
\end{figure}
\begin{figure}[h!]
    \begin{center}
        \includegraphics[width=0.95\textwidth]{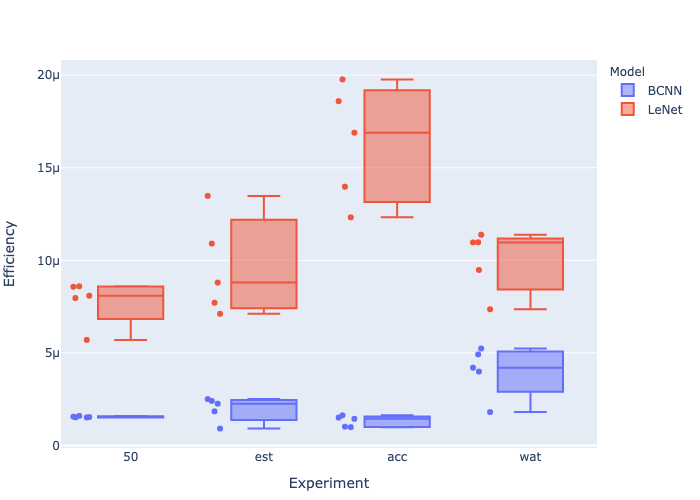}
    \end{center}
    \caption{Efficiency for the 4 experiments (CIFAR dataset).}\label{fig:cifar_efficency}
\end{figure}

Analyzing the relationship between stopping criteria and Efficiency in more detail, Figure~\ref{fig:mnist_efficiency} and Figure~\ref{fig:cifar_efficency} visually summarizes the efficiency analysis results from across the 50 epoch (50), early stopping (est), energy bound (wat), and accuracy bound (acc) experiments. In these figures, the x-axis indicates the stopping criteria of the models being assessed, the y-axis is the efficiency results per model size obtained from Equation~\ref{eq:efficiencybysize}, there are five model sizes for each architecture in each experiment, and so each box plot contains five efficiency results, and one box plot per stopping criteria.

Both Figure~\ref{fig:mnist_efficiency} and Figure~\ref{fig:cifar_efficency} show that different stopping criteria profoundly influence Efficiency. Variations in stopping criteria affect both the width of the distributions of efficiencies for each architecture and also the distance between these distributions. For example, stopping criteria that bound energy---the energy bound (wat) and the 50 epoch experiments---appear to squash the distributions of the model efficiencies of each architecture, whereas stopping criteria based on accuracy bounds---the early stopping (est) and accuracy bound (acc) experiments---the efficiency distributions are wider, particularly for the LeNet model. This energy bound versus accuracy bound categorization of stopping criteria is also predictive in terms of the gap between the LeNet and BCNN distributions, with accuracy bound experiments (est and acc) exhibiting a more significant gap between the distributions for the architectures as compared with the energy bound (wat and 50 epoch) experiments.

This suggests a trade-off between these two categories of stopping criteria for measuring architecture efficiency. Energy-bound experiments generate narrow efficiency distributions across model sizes, resulting in narrow confidence intervals around the mean Efficiency for a given architecture based on these experiments. However, this relatively more robust confidence is offset by the smaller gap between the efficiency distribution for each architecture. By contrast, the accuracy-bound experiments are more sensitive to differences between architectures in terms of Efficiency. However, the broader distribution per architecture results in wider confidence intervals around the mean Efficiency. In order to balance this trade-off, we suggest using both types of stopping criteria when measuring Efficiency (as done by Equation~\ref{eq:meaneffiencybyconvergence}).

The squashing of the distributions when energy-bound stopping criteria are used suggests that for each architecture, a fixed amount of energy per unit of accuracy is obtained independent of model size. In other words, when the stopping criteria are bound to energy, varying model size will not impact the overall architecture efficiency on a learning task. However, when the stopping criteria are based on accuracy, varying the model size will significantly impact the overall architecture efficiency of a learning task.

\begin{figure}[h!]
    \begin{center}
        \includegraphics[width=0.95\textwidth]{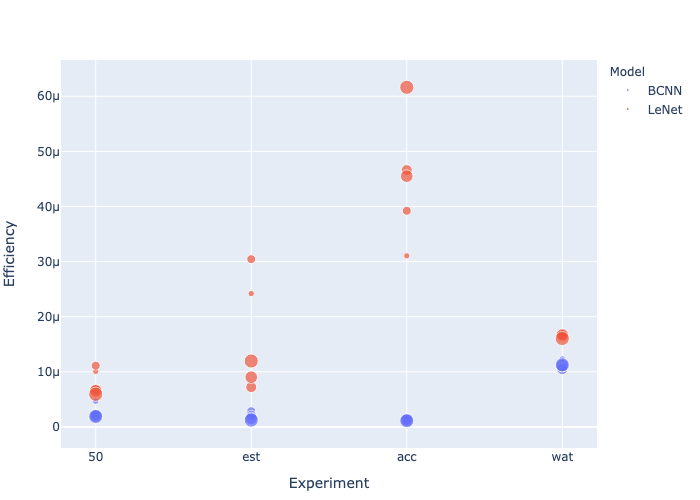}
    \end{center}
    \caption{Efficiency for the 4 experiments (MNIST dataset).}\label{fig:mnist_efficiency_sca}
\end{figure}

\begin{figure}[h!]
    \begin{center}
        \includegraphics[width=0.95\textwidth]{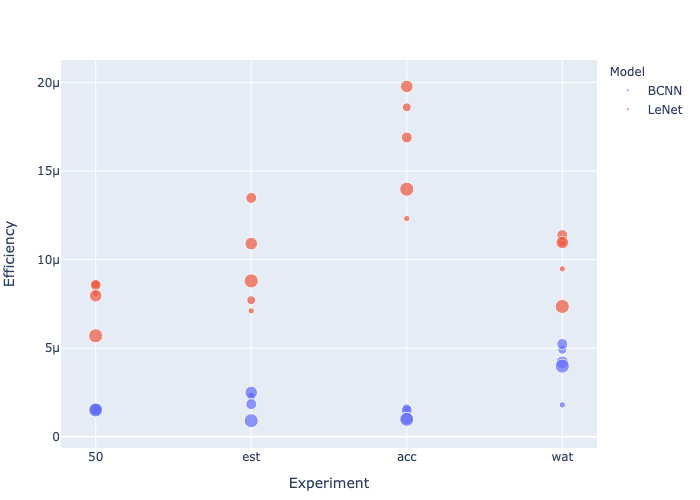}
    \end{center}
    \caption{Efficiency for the 4 experiments (CIFAR dataset).}\label{fig:cifar_efficiency_sca}
\end{figure}

Figure~\ref{fig:mnist_efficiency_sca} and Figure~\ref{fig:cifar_efficiency_sca} show how different model sizes compare to each other. The Efficiency of the LeNet architecture is particularly sensitive to size. However, there is no clear trend between size and Efficiency. The Efficiency of the BCNN architecture is less sensitive to size variation, so the visualizations are less helpful for this architecture. However, examining efficiency results reported in Tables~\ref{tab:mnist_100_eff},~\ref{tab:mnist_es_eff},~\ref{tab:mnist_eb_eff},~\ref{tab:mnist_ab_eff}, and Tables~\ref{tab:cifar_100_eff},~\ref{tab:cifar_es_eff},~\ref{tab:cifar_eb_eff},~\ref{tab:cifar_ab_eff} reveals that there is no apparent trend between model size and Efficiency.

\subsection{Efficiency of the LeNet architecture against BCNN architecture}\label{ssec:effbcnnvlenet}

Table~\ref{tab:results_global} presents the overall efficiency calculations for the LeNet and BCNN architectures on the MNIST and CIFAR datasets. These efficiencies are the mean Efficiency of architecture on a dataset across the multiple model sizes and convergence criteria (see Equation~\ref{eq:meaneffiencybyconvergence}). On both datasets, LeNet is more efficient than the BCNN architecture.

\begin{table}[h!]
    \centering
    \caption{Efficiency ($\eff(arch, convergence)$) of BCNN and LeNet architectures on the MNIST and CIFAR datasets.}
    \label{tab:results_global}
     \begin{tabular}{l c c c }
        \hline
        ~	& MNIST & CIFAR & MNIST/CIFAR\\
        \hline
        LeNet 	& $12.86 \times 10^{-6}$ & $7.06 \times 10^{-6}$ & $1.82$\\
        BCNN	& $2.88 \times 10^{-6}$ & $1.36 \times 10^{-6}$ & $2.11$\\
        LeNet/BCNN 	& $4.46$ & $5.18$ & ~ \\
        \hline
    \end{tabular}

\end{table}

Comparing the Efficiency of each architecture across the datasets, we can see that both architectures are more efficient on MNIST than on CIFAR. This is due to the relative complexity of CIFAR versus MNIST. In order to check how the Efficiency of an architecture varies across datasets, we take the ratio between the Efficiency of an architecture on one dataset to its Efficiency on another. In this calculation, we take Efficiency on MNIST as the numerator because this is the dataset on which both architectures have the highest Efficiency. For LeNet, this calculation is $12.86/7.06 = 1.822$, i.e., the LeNet architecture is $1.822$ times more efficient on MNIST than on  CIFAR. For BCNN, this calculation gives us $2.88/1.36 = 2.116$, i.e., the BCNN architecture is $2.116$ times more efficient on MNIST than CIFAR. These two ratios are close. However, the ratio for LeNet is smaller than for BCNN $1.822<2.116$, indicating that the LeNet architecture has a smaller decrease in Efficiency between MNIST and CIFAR than BCNN. Another perspective on these results is to take the ratio between the two architectures on each dataset. In this case, we use the Efficiency of the LeNet architecture as the numerator because this architecture has the highest Efficiency on both datasets. For MNIST, this calculation is $12.86/2.88 = 4.466$, and for CIFAR, this calculation is $7.06/1.36 = 5.185$. These calculations indicate that on MNIST, LeNet is $4.466$ times more efficient than BCNN, whereas on CIFAR, LeNet is $5.185$ times more efficient than BCNN. In other words, as the dataset becomes more complex (moving from MNIST to CIFAR), the difference in Efficiency between LeNet and BCNN becomes larger ($5.185>4.466$).

To summarise, the CIFAR dataset is the more complex dataset, LeNet is the more efficient architecture on both datasets, and when the learning task switches to a more complex dataset, the relative differences in Efficiency between the architectures become more extensive (the less efficient architecture has a more considerable relative drop in Efficiency, and the ratio between the efficiencies of the architectures increases as the task becomes more difficult). This observation relating the difficulty of the task and changes in the Efficiency of an architecture aligns with what can be observed in Figure~\ref{fig:mnist_eff_epo} and Figure~\ref{fig:cifar_eff_epo} where there is a more significant gap between the LeNet and BCNN plots lines on CIFAR as compared to the plot lines on MNIST.

This comparison of Bayesian Convolutional Neural Networks (BCNNs) and Convolutional Neural Networks (CNNs) highlights a trade-off in training efficiency. BCNNs seek to enhance generalizability by learning a distribution over models rather than fitting a single model to the data (thereby reducing the risk of overfitting)~\cite{mackay_bayesian_1995}. However, learning this distribution requires the repeated sampling of weights during training, which incurs an extra cost in terms of energy. For BCNNs to achieve greater Efficiency than CNNs, their generalization improvement must outweigh the increased energy costs incurred during training. Our findings indicate that, for the tasks we have examined, this trade-off results in BCNNs being less efficient than CNNs in terms of accuracy versus energy.

\subsection{On the risks of over-training (over-fitting)}
\label{subsec:overtraining}

As discussed in Section~\ref{secsub:efficiencyacrosstraining}, the Efficiency of a neural architecture tends to decay as training progresses; this trend is evident in Figure~\ref{fig:mnist_eff_epo} and Figure~\ref{fig:cifar_eff_epo} where for both architectures on both datasets efficiency consistently reduces as training progresses.
This trend reflects that as training progresses, model performance saturates after a certain point, and further training expends more energy with no gain in performance. An implication of this is that if a neural model is trained for an extreme number of epochs, then the training efficiency of that architecture will tend to zero, and furthermore, in such a scenario, comparing the Efficiency of different neural architectures is no longer sensible because all architectures will have an efficiency of zero. Put another way, the measurement of the training efficiency for a neural architecture only makes sense when models are not overtrained.

The most direct definition of overtraining is epochs of training that do not improve model performance. Another complementary way of identifying when overtraining has occurred is through the concept of over-fitting. Overfitting occurs when a model learns to perform well on the training data but fails to generalize to unseen data, compromising its Efficiency. Overfitting can be checked for by comparing the divergence between a model's performance on training data versus non-training data. To illustrate both overfitting and the impact of overtraining training efficiency, we extend our 50-epoch experiment to 100 epochs. We then perform two levels of analysis. First, we check whether the models trained for 100 epochs exhibit overtraining (compared to those trained for 50 epochs). Then, we calculate the Efficiency of both architectures using the results from the 100-epoch experiment in order to understand how overtraining can affect training efficiency.

We examine two measures to check whether extending training from 50 to 100 results in overtraining a model. First, we check whether the extra training resulted in an appreciable increase in model performance on the test set; if there is no increase in test set performance between the 50th and 100th epoch, then we deem the 100 epoch model to be overtrained. Second, suppose a model exhibits an increase in test set performance between the 50th and 100th epochs. We check for overfitting by comparing the model's performance on the training data and the test set. The intuition behind this analysis is that the more significant the drop in the performance between the training data and a test set, the more likely the model will be overfitted (and hence overtrained). In more detail, we calculate the difference between a model's training and test performance after 50 epochs of training and after 100 epochs of training and then calculate the delta between these differences. This delta in the differences reveals the extent of divergence between training and test performance caused by the extra 50 epochs of training. Using this delta metric, we deem a model to be overtrained if the delta is of a comparable scale to the increase in the test set performance of the model between the 50th and 100th epochs.

Table~\ref{tab:overfit} presents the performance results used in this analysis. For the 50 and 100 epoch results, the table presents the model performance on the training set, the test set, and the difference between these results. The rightmost two columns of the table (columns A and B) list the difference in test performance between 50 and 100 epochs (calculated as test performance at 100 epochs minus test performance at 50 epochs) and the delta in the differences between training and test performance between 50 and 100 epochs (calculated as the difference between training and test performance at 100 epochs minus the difference between training and test performance at 50 epochs). In order to highlight meaningful differences in columns A and B, we round the results in these columns to two decimal places. If we examine column A, we see that on the MNIST dataset, none of the LeNet models obtain a meaningful increase in test performance between the 50th and 100th epochs. As a result, we consider the LeNet 100 epoch models to be overtrained. The BCNN models on MNIST exhibit a slight increase ($\approx 0.01$ for all models) in test set performance between the 50th and 100th epoch. However, this is accompanied by a comparable increase in the divergence between training and test set performance, so we also deem these BCNN models to be overtrained. Switching focus to the CIFAR dataset, all of the LeNet models exhibit an increase in test performance between the 50th and 100th epoch. However, this is accompanied by a comparable (and in 4 out of 5 cases more prominent) increase in divergence between training and test performance, so we deem these 100 epoch LeNet CIFAR models to be overtrained. Finally, the BCNN models on the CIFAR dataset all exhibit a relatively significant increase in test performance between the 50th and 100th epoch, accompanied by a comparably slight increase in divergence between training and test performance, so we deem these models not to be overtrained. In summary, our analysis of overtraining after 100 epochs categorized all the LeNet and BCNN MNIST models, the LeNet CIFAR models as overtrained, and the BCNN CIFAR models as not overtrained.

\begin{table}[h!]
    \centering
    \caption{An analysis of model over-fitting after 100 epochs. Column A lists the per-model increase in test set performance between 50 and 100 epochs (Test accuracy after 100 epochs minus Test accuracy after 50 epochs). Column B lists the per model delta in the training and test difference between 50 and 100 epochs (Difference at 100 minus Difference at 50)}
    \label{tab:overfit}
     \rotatebox{90}{
     \begin{minipage}{0.45\textwidth}
    \begin{tabular}{ l l l l l l l c c}
    \toprule
        ~ & \multicolumn{3}{c}{50 epoch} & \multicolumn{3}{c}{100 epoch}& ~ & ~\\
        \midrule
        MNIST & Train & Test & Difference & Train & Test & Difference &  A & B \\
        \midrule
        LeNet-1 & 0.99071102 & 0.98506103 & 0.00564999 &  0.99524102 & 0.98694648 & 0.00829454 &  0.00 & 0.00 \\ 
        LeNet-2 & 0.99438373 & 0.98787305 & 0.00651068 &  0.99708797 & 0.98906119 & 0.00802678 &  0.00 & 0.00 \\ 
        LeNet-3 & 0.99525017 & 0.99059342 & 0.00465675 &  0.99761469 & 0.99154647 & 0.00606822 &  0.00 & 0.00 \\ 
        LeNet-4 & 0.99565284 & 0.99153448 & 0.00411836 &  0.99782642 & 0.99276667 & 0.00505975 &  0.00 & 0.00 \\ 
        LeNet-5 & 0.99629737 & 0.99057988 & 0.00571749 &  0.99814869 & 0.99128859 & 0.00686010 &  0.00 & 0.00 \\
        \midrule
        BCNN-1 & 0.95748047 & 0.96715924 & -0.00967877 &  0.97338431 & 0.97560624 & -0.00222193 &  0.01 & 0.01 \\ 
        BCNN-2 & 0.96626704 & 0.97242924 & -0.0061622 &  0.97819731 & 0.97928500 & -0.00108769 &  0.01 & 0.01 \\ 
        BCNN-3 & 0.96513256 & 0.97038611 & -0.00525355 &  0.97763173 & 0.97727584 & 0.00035589 &  0.01 & 0.01 \\ 
        BCNN-4 & 0.96491689 & 0.96950655 & -0.00458966 &  0.97758249 & 0.97698637 & 0.00059612 &  0.01 & 0.01 \\ 
        BCNN-5 & 0.96142537 & 0.96888013 & -0.00745476 &  0.97537774 & 0.97666603 & -0.00128829 &  0.01 & 0.01 \\
        \midrule
        CIFAR & Training & Testing & Difference &  Training & Testing & Difference & ~ & ~ \\
        \midrule
        LeNet-1 & 0.59485669 & 0.57332227 & 0.02153442 &  0.65716212 & 0.61330469 & 0.04385743 &  0.04 & 0.02 \\ 
        LeNet-2 & 0.70195860 & 0.64035352 & 0.06160508 &  0.76755101 & 0.66876953 & 0.09878148 &  0.03 & 0.04 \\ 
        LeNet-3 & 0.76978155 & 0.68373047 & 0.08605108 &  0.83391819 & 0.70827051 & 0.12564768 &  0.02 & 0.04 \\ 
        LeNet-4 & 0.80343700 & 0.69496875 & 0.10846825 &  0.86194093 & 0.71430078 & 0.14764015 &  0.02 & 0.04 \\ 
        LeNet-5 & 0.82096686 & 0.69295508 & 0.12801178 &  0.88013734 & 0.71263281 & 0.16750453 &  0.02 & 0.04 \\
        \midrule
        BCNN-1 & 0.33502488 & 0.33920898 & -0.0041841 &  0.43809589 & 0.43243262 & 0.00566327 &  0.09 & 0.01 \\ 
        BCNN-2 & 0.44060957 & 0.43411328 & 0.00649629 &  0.50035355 & 0.48572559 & 0.01462796 &  0.05 & 0.01 \\ 
        BCNN-3 & 0.45429289 & 0.44414062 & 0.01015227 &  0.52243183 & 0.50286328 & 0.01956855 &  0.06 & 0.01 \\ 
        BCNN-4 & 0.45937550 & 0.45482617 & 0.00454933 &  0.53284086 & 0.51709375 & 0.01574711 &  0.06 & 0.01 \\ 
        BCNN-5 & 0.46070014 & 0.45722070 & 0.00347944 &  0.53233031 & 0.51748633 & 0.01484398 &  0.06 & 0.01 \\
        \bottomrule
    \end{tabular}
     \end{minipage}}
\end{table}

To analyze how overtraining can affect the measurement of training efficiency, we used Equation~\ref{eq:efficiencybysize} to calculate the Efficiency of both architectures on both datasets based solely on the results of the 100 epoch experiment. The results of these calculations are presented in Table~\ref{tab:results_global_100}. We are comparing these results with those listed in Table~\ref{tab:results_global}; a consistent finding across both sets of results is that LeNet is more efficient than BCNN on both datasets. Also, for three out of the four categories of models (LeNet and BCNN on MNIST, and LeNet on CIFAR), the training efficiency drops as compared with Table~\ref{tab:results_global}, this is in line with what would be expected from the trends exhibited in Figure~\ref{fig:mnist_eff_epo} and Figure~\ref{fig:cifar_eff_epo} discussed in Section~\ref{secsub:efficiencyacrosstraining}. The one exception to this trend is the BCNN architecture on CIFAR, which slightly increases Efficiency. This exception aligns with the findings of our overtraining analysis presented above. It suggests that if we were to use the efficiency scores presented in Table~\ref{tab:results_global_100} to compare the efficiency scores of LeNet and BCNN, we would be comparing overtrained LeNet models against BCNN models, some of which are overtrained (i.e., BCNN MNIST) and some of which are not (i.e., BCNN CIFAR). If we run this (incorrect) comparison through to see how overtraining can affect the overall analysis, we get very different conclusions from those we reached from analyzing Table~\ref{tab:results_global}. For example, let us compare the efficiency ratio for each architecture across the two datasets (i.e., MNIST/CIFAR). We see that in Table~\ref{tab:results_global_100} for LeNet, this ratio ($1.042$) is greater than the BCNN ratio ($0.367$).
Similarly, if we compare the efficiency ratio between the two architectures on each dataset (LeNet/BCNN), we see that this ratio is more significant for MNIST ($3.108$) than for CIFAR ($1.095$). In both cases, the relative size of these ratios has flipped as compared with the results reported in Table~\ref{tab:results_global}. Taking the ratios in Table~\ref{tab:results_global_100} at face value, we would (erroneously) conclude that as the learning task becomes more complex (MNIST$\rightarrow$CIFAR), the more efficient architecture (LeNet) has a more significant drop in Efficiency and that the difference in Efficiency between the two architectures becomes smaller. However, the underlying phenomenon driving these results is overtraining. Consequently, when assessing the training efficiency of neural architecture, it is essential to consider overtraining as a factor in the analysis and to be cognizant that overtraining can occur at different points in training for different models on a given training task. One strategy to mitigate the risk of overtraining impacting efficiency analysis is to average over multiple convergence criteria, as we have done in this work.

\begin{table}[h!]
    \centering
    \caption{Efficiency ($\eff(arch, convergence)$) of BCNN and LeNet architectures on the MNIST and CIFAR datasets for models trained for 100 epochs.}
    \label{tab:results_global_100}
     \begin{tabular}{l c c c }
        \hline
        ~	& MNIST & CIFAR & MNIST/CIFAR\\
        \hline
        LeNet 	& $2.05 \times 10^{-6}$ & $1.97 \times 10^{-6}$ & $1.04$\\
        BCNN	& $0.66 \times 10^{-6}$ &  $1.80 \times 10^{-6}$ & $0.36$\\
        LeNet/BCNN 	& $3.10$ & $1.09$ & ~ \\
        \hline
    \end{tabular}

\end{table}

\section{Conclusions}\label{sec:conclusion}

We present a framework for measuring the training efficiency of a neural architecture on a learning task. This framework involves running multiple experiments but does not require hardware profiling. Moreover, the framework enables a multifaceted analysis of the training efficiency of a neural architecture, including the analysis of how the Efficiency of a model varies across training epochs (Equation~\ref{eq:efficiencyvalidated}), how the Efficiency of a neural architecture varies with model size (Equation~\ref{eq:efficiencybysize}) and the overall Efficiency of a neural architecture on a learning task taking into account variations in model size and stopping criteria (Equation~\ref{eq:meaneffiencybyconvergence}). Furthermore, the ability to calculate an overall efficiency for a neural architecture on a learning task enables the analysis of the relative Efficiency of different neural architectures on a learning task and how the relative Efficiency of neural architectures varies across learning tasks.

Applying the framework to the case study comparing CNNs with BCNNs on MNIST and CIFAR, we find that the Efficiency of both architectures on both learning tasks changes substantially as training progresses (see Section~\ref{secsub:efficiencyacrosstraining}), with all models exhibiting a drop in Efficiency across epochs. The analysis in Section~\ref{ssec:sizeandefficiency} reveals a non-linear relationship between stopping criteria and training Efficiency and model size and training Efficiency. We observed significant variation in training efficiency across different stopping criteria for both architectures. This variation across stopping criteria illustrates the need for multiple stopping criteria within the efficiency framework. Moreover, including multiple convergence criteria within the framework mitigates the risk of overtraining affecting the analysis of the training efficiency of neural architectures (see Section~\ref{subsec:overtraining}). More generally, we believe that the potential confounding effect of overtraining on neural training efficiency research is not given sufficient attention in the literature. To take a recent example, \cite{KaddourEtAl-2023} report, as a key finding, that the efficiency improvements obtained by several training regime modifications vanished when the compute budget allowed for training increases. However, in their analysis, the authors did not consider that this finding may result from overtraining occurring at different points under different training regimes. Indeed, the more efficient a training regime is, the earlier in the training process overtraining will begin, in which case, using a fixed compute budget as a convergence criterion is likely to result in more efficient training regimes overtraining for longer. So, the extra overtraining will negate the efficiency benefits of these regimes. This example illustrates how neglecting the impact of overtraining can directly undermine conclusions drawn from an experiment focused on training efficiency. Regarding the relationship between model size and training Efficiency, we find that intermediate-size models have the best Efficiency for both architectures and learning tasks. This variation in Efficiency with respect to model size highlights the need to include model size within the efficiency frameworks.

In terms of overall neural architecture training efficiency on a learning task, we find that CNNs are more efficient than BCNNs on both MNIST and CIFAR and that the difference in Efficiency becomes more prominent as the learning task becomes more complex (see Section~\ref{ssec:effbcnnvlenet}). To test for interactions with hardware, we replicated our experiments and analysis on a second hardware setup. The description of the hardware and the results are presented in~\ref{apx:hardware_results}. The same trends are evident in the results obtained from these other experiments. Overall, we argue that to measure the training efficiency of neural architectures, it is important to consider efficiency variation across model size, the stopping criterion used, and the learning task. In future work, we will explore the application of the framework to other neural architectures and training paradigms. For example, there is a growing body of work exploring parameter-efficient fine-tuning, and applying this framework to these methods could reveal important interactions between the neural architecture and the training regimen. Another potential area of future work emerges from our findings that training efficiency and model size have a non-linear relationship. Given this finding, it may be helpful to consider how Efficiency, model size, and model compression methods interact\footnote{Supplementary material is available at the~\href{https://osf.io/qw7rj/?view_only=56f77a9e6a8245048b2531c9d3a076b0}{Open Science Foundation}.}.

\begin{appendices}

\section{Hardware comparison}\label{apx:hardware_results}

We replicated our experiments on a second hardware setup to demonstrate our framework's generalizability and findings.
Table~\ref{tab:hardware_char} shows the characteristics of this second (AMD) hardware platform. Due to the smaller capabilities of this hardware platform, the training regime was modified for the CIFAR dataset; instead of using six-fold validation, we used a single 70-30 split on the data. This modification allows the training to be completed on this AMD hardware without any memory overflow.
Apart from this modification, the same training regimen, architectures, and hyperparameters as described in Section~\ref{sec:casestudy} were used in these experiments.

\begin{table}[h!]
    \caption{Hardware characteristics.}
    \label{tab:hardware_char}
    \begin{tabular}{l}
    AlmaLinux 9.2 (Turquoise Kodkod) x86\_64 \\
    Kernel: 5.14.0\-284.11.1.el9\_2.x86\_64  \\
    CPU: AMD Ryzen 9 5900HX with Radeon Graphics (16) @ 3.300GHz \\
    GPU: AMD ATI Radeon Vega Series / Radeon Vega Mobile Series \\
    GPU: AMD ATI Radeon RX 6700/6700 XT/6750 XT/6800M/6850M XT \\
    Memory: 3251 MiB / 31496 MiB \\
    Driver version: 6.1.5 \\
    ROCm version: 5.4.2 \\
    Python version: 3.9.16 \\
    Pytorch version: 2.0.1 \\
    powerstat version: 0.03.03 \\
    radeontop version: 1.00 \\
    \end{tabular}
\end{table}

The experimental data was processed in the same manner as in Section~\ref{sec:measure_results}, obtaining the following results:

\begin{figure}[H]
    \begin{center}
        \includegraphics[width=0.55\textwidth]{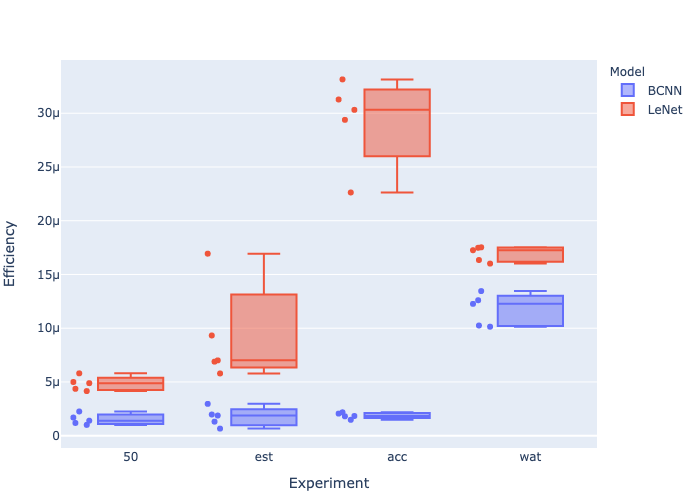}
    \end{center}
    \caption{Box plot for Efficiency per size four experiments (MNIST dataset).}\label{fig:amd_mnist_efficiency_box}
\end{figure}

\begin{figure}[H]
    \begin{center}
        \includegraphics[width=0.55\textwidth]{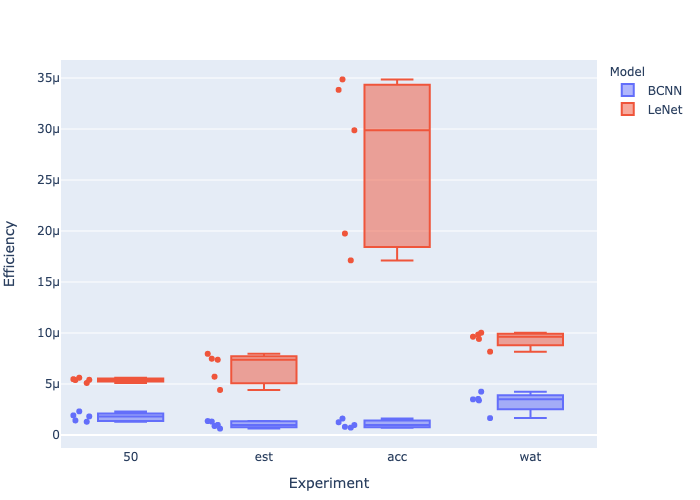}
    \end{center}
    \caption{Box plot for Efficiency per size four experiments (CIFAR dataset).}\label{fig:amd_cifar_efficiency_box}
\end{figure}

\begin{figure}[H]
    \begin{center}
        \includegraphics[width=0.55\textwidth]{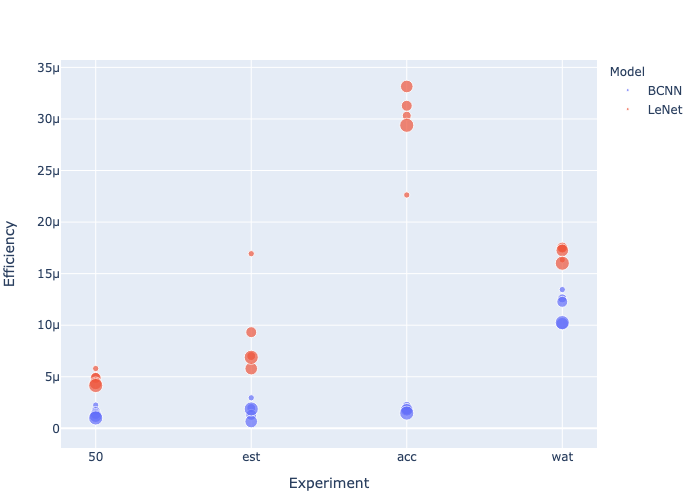}
    \end{center}
    \caption{Scatter plot for the Efficiency 4 experiments (MNIST dataset).}\label{fig:amd_mnist_efficiency_scat}
\end{figure}

\begin{figure}[H]
    \begin{center}
        \includegraphics[width=0.55\textwidth]{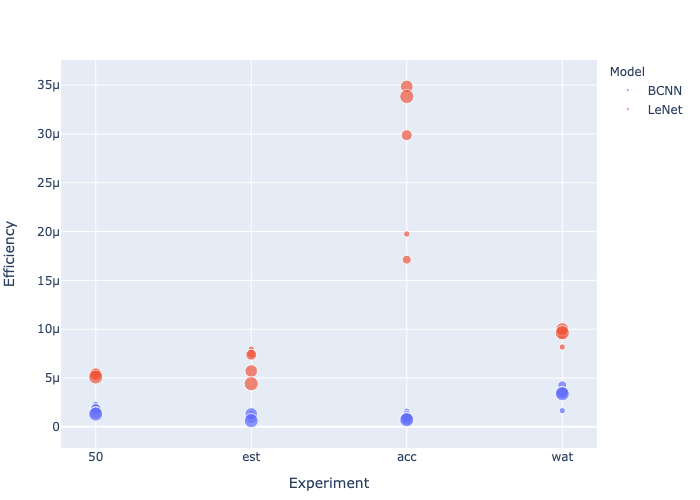}
    \end{center}
    \caption{Scatter plot for the Efficiency 4 experiments (CIFAR dataset).}\label{fig:amd_cifar_efficiency_scat}
\end{figure}

The results from the data collected are similar to the ones presented in Section~\ref{sec:analysingflops}.

\begin{table}[h!]
    \centering
    \caption{Efficiency ($\eff(arch, convergence)$) of BCNN and LeNet architectures on the MNIST and CIFAR datasets, with AMD hardware.}
    \label{tab:results_global_amd}
     \begin{tabular}{l c c c }
        \hline
        ~	& MNIST & CIFAR & MNIST/CIFAR\\
        \hline
        LeNet 	& $8.91 \times 10^{-6}$ & $19.30 \times 10^{-6}$ & $0.46$\\
        BCNN	& $2.66 \times 10^{-6}$ & $1.18 \times 10^{-6}$ & $2.25$\\
        LeNet/BCNN 	& $3.35$ & $16.41$ & ~ \\
        \hline
    \end{tabular}

\end{table}


Table~\ref{tab:results_global_amd} shows that our results over the MNIST dataset and CIFAR dataset, for both neural architectures, across both hardware manufacturers seem consistent, i.e., they follow a similar trend and clearly show that the LeNet architecture is more efficient overall than the BCNN architecture, similar to Section~\ref{ssec:effbcnnvlenet}.

Figures~\ref{fig:amd_mnist_efficiency_box} and Figure~\ref{fig:amd_cifar_efficiency_box} follow along the analysis presented in Section~\ref{ssec:sizeandefficiency}, with Figure~\ref{fig:amd_mnist_efficiency_scat} and Figure~\ref{fig:amd_cifar_efficiency_scat}, following a similar trend. These results validate that the Efficiency reported and the analysis presented are consistent across hardware platforms.\\

\textbf{Acknowledgements} This work was conducted with the financial support of the Science Foundation Ireland Centre for Research Training in Digitally-Enhanced Reality (d-real) under Grant No. 18/CRT/6224.\\

\section*{Declarations}

\textbf{Competing interests} The authors declare no competing interests.

\textbf{Open Access} For the purpose of Open Access, the author has applied a CC BY public copyright license to any Author Accepted Manuscript version arising from this submission.




\end{appendices}


\bibliography{sn-bibliography}

\end{document}